# Large-Scale Video Search with Efficient Temporal Voting Structure

Ersin Esen, Savaş Özkan, İlkay Atıl
Image Processing Group
TÜBİTAK UZAY, Ankara, TURKEY
{ersin.esen, savas.ozkan, ilkay.atil}@tubitak.gov.tr

*Abstract*— In this work, we propose a fast content-based video querying system for large-scale video search. The proposed system is distinguished from similar works with two major contributions. First contribution is superiority of joint usage of repeated content representation and efficient hashing mechanisms. Repeated content representation is utilized with a simple yet robust feature, which is based on edge energy of frames. Each of the representation is converted into hash code with *Hamming Embedding* method for further queries. Second contribution is novel queue-based voting scheme that leads to modest memory requirements with gradual memory allocation capability, contrary to complete brute-force temporal voting schemes. This aspect enables us to make queries on large video databases conveniently, even on commodity computers with limited memory capacity. Our results show that the system can respond to video queries on a large video database with fast query times, high recall rate and very low memory and disk requirements.

## I. Introduction

Modern content-based multimedia indexing systems are faced with an explosion in the amount of multimedia data due to the widespread use of mobile devices and increasing popularity of multimedia content on social media. The recent studies[1] have estimated that the %80 of total internet traffic will be used for video transmission by 2019. Unfortunately, current video search methods are not completely capable of handling such a huge amount of video archives within reasonable computation times for both content representation and querying extents.

In literature, most of the works related to video search task are based on image-based search methods due to the effectiveness against severe geometric distortions, compression, color changes etc. [1, 2, 9-11, 13]. In these methods, content of a video is analyzed on some of the sampled frames and sequential ordering is imposed lately during pairing process of frame corresponds in both query and reference videos. This induces a weakness since temporal content model is partially discarded in content representation. In [3], the authors assert that using both spatial and temporal content models as in activity recognition can increase the performance of video content search while spatial-temporal consistency between matching pairs is still preserved.

Usage of frames in content representation can generate huge burdens on storing space and computational cost as video databases grow. This aspect of frame-based video representation greatly limits the scalability of video querying systems. To overcome this problem, different works in the literature are proposed in which more simplified video representation are utilized [4, 12]. In these methods, robustness is preserved up to some degree whilst the main objective is to achieve fast content representation. [4] computes sparse extrema points from one dimensional histogram which is estimated with global motion changes in consecutive frames. The authors show that these locations are sufficiently robust to several distortions and similar characteristic can be still obtained on the attacked version of the video. Even though frequency domain responses around these points provide distinct information for accurate pair matching, such simple features become more effective when number of paired points in time is increased. This forms the underlying idea behind the repeated content representation scheme.

Second critical deficiency in video search is that in the retrieval process of similar correspondences from a database, temporal consistencies within paired videos should be preserved with either matching frames or the assumption discussed above. To realize this feature, large tables must be generated. This essentially causes huge memory and computation needs and it limits the scalability of temporal matching on very big video databases. If we consider that people watch hundreds of millions hours of video on YouTube[2] every day, it is nearly impossible to handle this amount with such system. That's why, wise approaches need to be proposed for more effective search mechanism.

In this work, we propose a novel video querying system which is fast, has low memory and storage space requirements. We can summarize our novelties as follows:

- Using *"Edge Energy"* (EE) feature to represent a repeated content of a video. This is similar to [4] but it is cheaper to compute. Furthermore, high dimension space is alleviated with a hashing method described in [1] to store video representations in compact forms.

---

[1] Cisco Visual Networking Index: Forecast and Methodology, 2014–2019
[2] https://www.youtube.com/yt/press/statistics.html

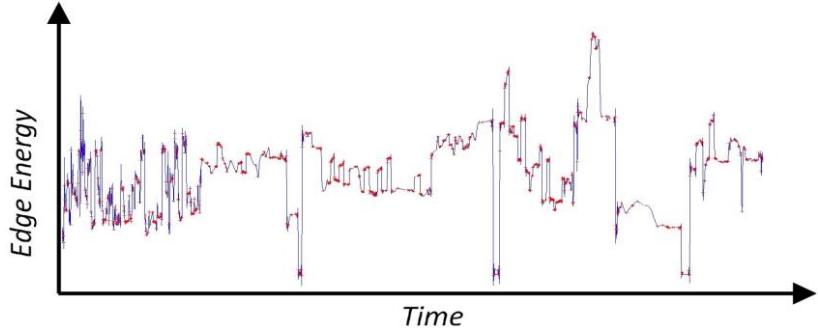

Figure 1: *Edge Energy Change (EEC)* feature characteristic for a video. Red points on the plot indicate local extrema points where sparse content representations for a video are computed.

- Using novel adaptive temporal voting scheme to keep sequential matching in small memory range compared to complete brute-force voting table methods. This greatly scales the space complexity and extend larger-scale video search capability.

The paper is organized as follows. Section II explains the details of the proposed method. Section III includes the experimental results conducted on a domain-adapted test video dataset and the discussions in light of the test results. Section IV summarizes our work and emphasizes strong/weak points of the proposed system.

## II. PROPOSED METHOD

In this section, we describe the overall structure of our proposed system. Therefore, first, we give a brief explanation of the system and then we detail the design of processing steps.

Our system mainly consists of two steps: Video content representation and video querying. In the first step, we extract *Edge Energy* (EE) features and compute projections of these features onto hash codes with [1] to represent content of a video with robust signatures. In the second step, similar video segments including part(s) of a query video are retrieved with a novel queue-based temporal voting scheme. Matching scores between signatures are computed by comparing the hash codes on both query and reference videos as described in previous step. For the sake of clarity, we provide a notation table in Table I. It holds brief definitions of the notations and their corresponding symbols.

### A. Video Content Representation

#### 1) Edge Energy (EE)

In [4], the authors show that content changes between consecutive frames exhibit sufficiently distinct characteristics about video content. With slight modification i.e. replacing motion change feature with more stable one, we completely stick to this assumption in this paper. In the flow of the representation, edge energy value on each video frame is computed and one dimensional histogram is formed by concatenation of these values in time series. The main improvement of edge energy feature over motion changes as in [4] is to enable to model content with faster processing capacity while achieving compatible performances.

Equation 1 shows the calculation of edge energy value for a single frame. Edge energy $e_t$ of a video frame at time $t$ is calculated by summing *Sobel* filter results of a frame on x and y directions and normalizing the summation with the frame size. When edge energy values are gathered for multiple frames of a video, we obtain a one dimensional *Edge Energy* (EE) feature (Figure 1) which can be used to represent a video in a robust and compact form.

$$e_t = \frac{1}{M} \sum_{k=1}^{M} \sqrt{g_x^{k^2} + g_y^{k^2}} \qquad (1)$$

TABLE I. NOTATION USED IN DEFINITIONS

| Symbol | Definition |
| --- | --- |
| $t$ | Time index of the video |
| $f_t$ | Video frame at time $t$ |
| $e_t$ | Edge energy of frame $t$ |
| $M$ | Number of pixels in the frame |
| $N_T$ | Hanning window side length |
| $N_F$ | Feature dimension |
| $v$ | Content representation for a frame |
| $N_C$ | Codebook dimension |
| $C$ | Codebook for feature space |
| $c_i$ | Codebook centroid |
| $q_v$ | Codebook index of the hash $v$ |
| $b_v$ | Binary code of the hash $v$ |
| $\tau_H^i$ | Threshold for $i^{th}$ dimension of the binary code |
| $\tau_{sc}$ | Similarity score threshold for binary code |
| $N_{nn}$ | Number of nearest neighbours |
| $seg$ | Video segment structure composed of video *id* and time $t$ |
| $q$ | Superscript denoting query |
| $r$ | Superscript denoting reference |
| $tol_{err}$ | Error toleration between consecutive frame numbers of a segment |
| $tol_{delete}$ | Temporal range for a segment to be considered as inactive |
| $n_{conf}$ | Minimum necessary number of matches for a segment to be considered as recurring |

where $g_x$, $g_y$ are the gradient values in the corresponding dimensions.

Selection procedure of ideal frame locations on the histogram is an important decision that can affect the performance of the retrieval significantly. Simply, selecting frames from fixed-intervals to compute content features is not desirable because of the fact that representation power of each frame might not be equal. Instead, we pick frames where the edge changes in time correspond to local extrema points. In this way, we consider only the frames which have distinct characteristics for video representation and matching.

To provide extra robustness, we apply *hanning* window around each local extrema by $2 \times N_T + 1$ window size with weighted sampling strategy. This basically gives more an emphasis around the center points and it gradually decreases weights for the samples close to window borders. Then frequency domain coefficients inside this window are calculated with Discrete Fourier Transform (DFT). We drop the DC coefficient and select the magnitude of the first $N_F$ DFT coefficients as the feature. By this assumption, we preserve most informative characteristics of the video content and eliminate non-discriminative ones.

*2) Hash Code*

Another advantage of sparse sampling *EE* features is that it drastically reduces the total number of representation and thus relieves the number of operations in comparison. However, even with this assumption, lots of high dimensional representations are computed. With basic approaches such as brute-force linear search, comparison of all *EE* representations between two videos will require impractical computation times for large video databases. Hence, we use a hashing method described in [1] to convert our EEC representations into compact signatures.

Each *EE* representation vector $v$ is first projected onto a pre-clustered feature space $C^{N_F \times N_C}$. Then closest cluster center $q_v$ and additional binary code $b_v$ are calculated by Equation 2 and 3 respectively. The dimension of a binary code is equal to $N_F$.

$$q_v = \arg\min_i \|v - c_i\|^2 \qquad (2)$$

$$b_v^{\ k} = \begin{cases} 1, & if\ v^k \geq \tau_H^k \\ 0, & otherwise \end{cases} \qquad (3)$$

Binary codes provide additional distinctions for the representations since usage of only the cluster center information might lead to a confusion in the separation of non-relevant features. As a result, we compare both $q_v$ and $b_v$ codes to produce the final decision about the similarity of two features. For the brevity, a representation $v_i$ computed on a frame will be considered as $q_v$ and $b_v$ codes from this point on.

To calculate the similarity of two representations $v_1$ and $v_2$, we consider relation between the cluster centers and binary codes jointly as in Equation 4:

$$sc(v_1, v_2) = \begin{cases} \sum_{k=0}^{N_F} \delta(b_{v_1}^k, b_{v_2}^k), & q_{v_1} == q_{v_2} \\ 0, & otherwise \end{cases} \qquad (4)$$

where $\delta(.)$ is the function that returns the total number of equal binary elements.

*B. Video Querying*

Different than image domain, in order to make proper decision about the relatedness of two videos, the distributions of similarity scores between individual pairs in time need to be consistent. This means that distributions should be concentrated on a single time instance with minimum saturation for true estimates. This prevents mismatches and improves the accuracy of a retrieval.

In order to keep track of multiple matches, a voting table need to be utilized and later true parts(s) of a video is determined based on the scores on this tables [6]. As expected from partial copy detection [13], this method works well and improves the success rate [2, 3]. However, one of the drawback of this kind of methods is that a large memory space should be allocated initially and thus this limits the scalability of a complete video search.

In this paper, we propose a novel queue-based voting scheme which keeps only small number of scores and gradually allocates memory according to needs. Inevitably, this greatly reduces the space requirement on matching process. Figure 2 shows the temporal voting mechanism proposed in this work. Simply, when a query video is given to our system, we first compute $\{q_v, b_v\}$ pairs from the video content as stated in the previous section. Then, for each $(q_v, b_v)$ pair, we retrieve $N_{nn}$ most similar video signatures from the reference video database using Equation 4. We store reference video signatures with an inverted index structure which provides faster accessibility during the code comparison. We then run a control mechanism that checks temporal association between each signature pair whether that is already presented in the voting queue or not by imposing time difference up to some error range as shown in Equation 5. In Equation 5, $seg_a$ and $seg_b$ respectively correspond to the segment for new pair and the segment that is already in the voting queue. If $seg_a$ is resolved to be too similar with $seg_b$, then we update $seg_b$ by increasing its vote by one and updating its start and end times. On the contrary, if $seg_a$ is determined as a different segment, then we append $seg_a$ into the voting queue as a new segment.

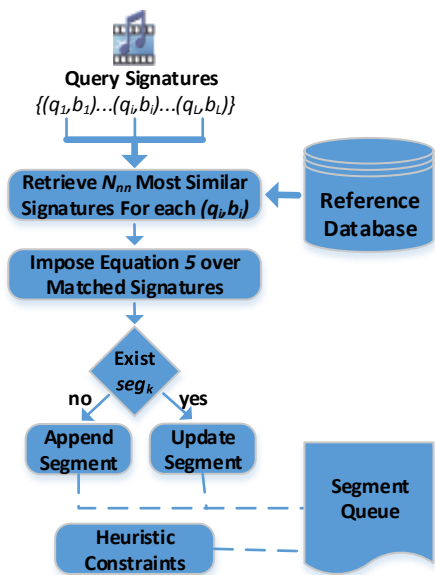

Figure 2: Temporal voting mechanism used to match hash codes for a given query video with all the other videos in a reference database.

TABLE II: MEAN AVERAGE PRECISIONS WITH AVERAGE QUERY TIME IN SECOND FOR DIFFERENT ATTACK TYPES ON 3800 HOURS DATASET

| 200 Kbps | | $N_T$ | |
|---|---|---|---|
| | | 75 | 100 |
| | 16 | 0,643 / 4,01 sec | 0,712 / 3,66 sec |
| $N_F$ | 32 | 0,916 / 2,67 sec | 0,916 / 2,49 sec |
| | 48 | 0,916 / 2,79 sec | 0,916 / 1,57 sec |

| 100 Kbps | | $N_T$ | |
|---|---|---|---|
| | | 75 | 100 |
| | 16 | 0,75 / 4,00 sec | 0,712 / 3,62 sec |
| $N_F$ | 32 | 0,916 / 2,74 sec | 0,916 / 2,74 sec |
| | 48 | 0,916 / 2,78 sec | 0,916 / 1,57 sec |

| 50 Kbps | | $N_T$ | |
|---|---|---|---|
| | | 75 | 100 |
| | 16 | 0,742 / 4,00 sec | 0,75 / 3,62 sec |
| $N_F$ | 32 | 0,916 / 2,80 sec | 0,916 / 2,49 sec |
| | 48 | 0,916 / 2,78 sec | 0,916 / 1,58 sec |

| 25 Kbps | | $N_T$ | |
|---|---|---|---|
| | | 75 | 100 |
| | 16 | 0,469 / 4,11 sec | 0,545 / 3,75 sec |
| $N_F$ | 32 | 0,75 / 2,82 sec | 0,833 / 2,51 sec |
| | 48 | 0,75 / 2,89 sec | 0,916 / 1,63 sec |

| 200 Kbps + logo | | $N_T$ | |
|---|---|---|---|
| | | 75 | 100 |
| | 16 | 0,166 / 3,87 sec | 0,083 / 3,56 sec |
| $N_F$ | 32 | 0,416 / 2,78 sec | 0,378 / 2,61 sec |
| | 48 | 0,469 / 2,89 sec | 0,5 / 1,70 sec |

| 100 Kbps + logo | | $N_T$ | |
|---|---|---|---|
| | | 75 | 100 |
| | 16 | 0,166 / 3,87 sec | 0,166 / 3,56 sec |
| $N_F$ | 32 | 0,371 / 2,79 sec | 0,378 / 2,64 sec |
| | 48 | 0,469 / 2,89 sec | 0,5 / 1,74 sec |

| 50 Kbps + logo | | $N_T$ | |
|---|---|---|---|
| | | 75 | 100 |
| | 16 | 0,166 / 3,64 sec | 0,166 / 3,38 sec |
| $N_F$ | 32 | 0,189 / 2,73 sec | 0,416 / 2,61 sec |
| | 48 | 0,416 / 2,83 sec | 0,5 / 1,75 sec |

| 25 Kbps + logo | | $N_T$ | |
|---|---|---|---|
| | | 75 | 100 |
| | 16 | 0,022 / 3,59 sec | 0,083 / 3,38 sec |
| $N_F$ | 32 | 0,25 / 2,84 sec | 0,25 / 2,60 sec |
| | 48 | 0,242 / 2,90 sec | 0,416 / 1,74 sec |

$$same(seg_a, seg_b) = \begin{cases} 1, & abs\big((t_a^q - t_a^r) - (t_b^q - t_b^r)\big) < tol_{err} \\ 0, & otherwise \end{cases} \quad (5)$$

Simultaneously, we control each segment in the voting queue periodically to remove defunct segments which have not received any votes for a certain time period $tol_{delete}$ from query segments. This ensures that the voting queue does not inflate too much with old segments which does not have a chance to get a vote through Equation 5. This also relieves the number of operations executed in segment comparison. At the final step of segment voting, we sort queue with respect to their votes, merge overlapping results and apply a vote threshold $n_{conf}$ to eliminate segments which has less votes than $n_{conf}$.

We should emphasize that since we retrieve top $N_{nn}$ similar signatures, retrieval process is not exact but approximate. However, missing some of the relevant features can be recovered by the tolerance we introduce in Equation 5.

III. EXPERIMENTS

For the system test, we downloaded 12 different commercial videos from YouTube whose durations vary from 30 seconds to 2 minutes[2]. These videos are combined in different sequential orders and 11 reference videos are generated. Furthermore, 3800 hours of distractor dataset [7] is inserted to reference database to simulate more realistic video search scenario and we ensure that none of the distractor videos contains part of query videos. For video query, we distorted original videos with severe re-encoding and logo insertion as illustrated in Figure 3.

We deliberately conduct our tests on a different dataset rather than well-known copy or partial duplicate datasets [13, 14] One of the main reason is that the deployed model for visual content representation needs regular content switching in time to compute more reliable features. However, in [13, 14], query durations are generally too short and thus our method does not find sufficient number of content features. That's why, we select commercial videos as queries which suit to our assumptions.

The configuration parameters of our system are determined in accordance with empirical results as follows: permitted maximum number of nearest neighbors $N_{nn}$ per signature is set to 200. The conducted tests verify that increasing the value of this parameter causes exponential slowdowns in query time. Conversely, decreasing the value discards true signature pairs and hence disrupts the precision rate. $tol_{err}$ and $n_{conf}$ are set to 2 and 200 respectively in terms of number of frames. For hash code generation, feature space is clustered into 4000 centroids using k-means algorithm and contribution of feature length $N_F$ is tested with varying numbers: 16, 32 or 48.

We provide mean average precision scores for some of the configuration parameters in Table II. Each element of the table correspond to mean average precision and average query time respectively. Query time is reported in seconds for 1 minute query on 3800 hours of reference videos. At least % 60 of query and reference videos need to be overlapped in time in order to consider retrieve score as 1.0. For final precision score, this procedure is repeated for all reference and query videos and average of them is set as final value.

---
[2] http://ersinesen.appspot.com/rekbul

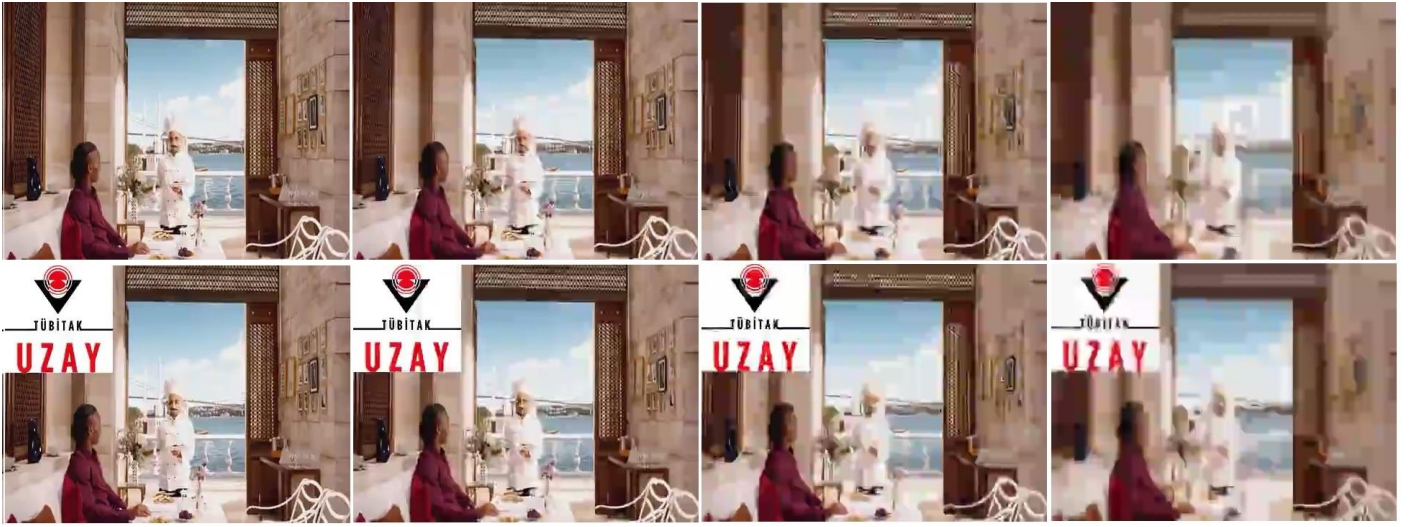

Figure 3: Example frames from a video query after different re-encoding and re-coding+logo attacks.

From the results, the proposed system achieves overwhelmingly fair results with higher $N_T$ and $N_F$ configurations for both re-encoding and logo distortions. The reason is that modelling video content with bigger window sizes provides robustness against visual distortions. Similarly, longer binary codes splits features space into more distinguishable partitions. Thus, following the assumption for larger dataset seems to be logical. For 0.916 precision rate in Table II, our system misses only one query video that consists of single shot. Some of the frames sampled sequentially are shown in Figure 4. Thus, it shows that sparse extrema points for *EE* features are mostly concentrated on visual shot boundaries [4].

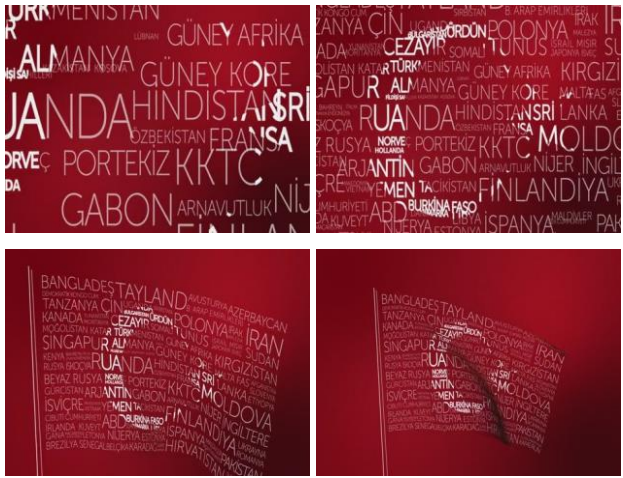

Fig. 4: Frame seqeunce from the query video that our approach misses.

Despite the high scores on re-encoding attacks, substantial performance drops can be observed for logo insertion. Even if *EE* features are robust to constant contrast modifications, *EE* feature and projected signature are altered due to the background changes around the inserted logo in time. The individual test results show that retrieval system works fine when distractor dataset is small or brute-force temporal table is used instead of our voting-queue. The limit in the number of nearest neighbors is the main reason behind this drop.

Another important result is of relieving memory burden. In our dataset, using brute-force method needs approximately 0.1 GB memory space to allocate the score tables. On the contrary, for our queue-based voting scheme, this drops down up to 0.01 GB which is impressive. Theoretically, we can emphasize that this memory requirement does not change even if dataset size increase since there is a limit on permitted number of nearest neighbors. However, due to the reason that we presented in previous paragraph, true pairs can be discarded excessively. Therefore, the balance should be adjusted well according to the application.

The query time is another important aspect for large-scale video search to evaluate the capacity of a system. The retrieval time in our system is really fast. Similarly, for higher $N_T$ and $N_F$ parameters, response time is getting smaller and becomes more practical to be used in a real video search system. Also, the whole database file size takes only 380 Mb which is convenient to be used in large-scale search application.

## IV. CONCLUSION

In this work, we propose a fast and efficient video content search system. The proposed system consists of two novelties. The first one is the fast video content representation based on edge energy feature and a hash code. *EE* feature is very fast to extract and requires very small storage, and hash code increases the scalability for both search and storage. The second and foremost contribution is the novel video temporal voting mechanism. This mechanism utilizes a moving vote accumulation over an inverted index feature database and significantly reduces required amount of memory with respect to conventional voting schemes.

The proposed system is tested with a reference video dataset of over 3800 hours for which ground truth of target video is available. The results indicate the high accuracy and efficiency of the proposed system.